\documentclass{article}
\usepackage{style_conf}

\usepackage{times}
\usepackage{soul}
\usepackage{url}
\usepackage[utf8]{inputenc}
\usepackage[small]{caption}
\usepackage{graphicx}
\usepackage{amsmath}
\usepackage{amsfonts}
\usepackage{amsthm}
\usepackage{booktabs}
\usepackage{algorithm}
\usepackage{algorithmic}
\usepackage[switch]{lineno}
\usepackage[colorlinks,linkcolor=blue,citecolor=blue]{hyperref}
\newcommand{\ie}{\textit{i.e.}}
\newcommand{\etal}{\textit{et al.}}

\title{Which Model to Transfer? A Survey on Transferability Estimation}

\author{
Yuhe Ding$^1$\and
Bo Jiang$^{1*}$\and
Aijing Yu$^{4,5}$\and
Aihua Zheng$^2$ \And
Jian Liang$^{3,5}$\footnote{Corresponding authors: Bo Jiang and Jian Liang.} \\
\affiliations
$^1$ School of Computer Science and Technology, Anhui University \\
$^2$ School of Artificial Intelligence, Anhui University \\
$^3$ CRIPAC \& MAIS, Institute of Automation, Chinese Academy of Sciences (CAS)\\
$^4$ Institute of Information Engineering, CAS $^5$ University of Chinese Academy of Sciences
\\ \texttt{madao3c@foxmail.com,jiangbo@ahu.edu.cn,liangjian92@gmail.com}
}

\begin{document}
\maketitle

\begin{abstract}
Transfer learning methods endeavor to leverage relevant knowledge from existing source pre-trained models or datasets to solve downstream target tasks.
With the increase in the scale and quantity of available pre-trained models nowadays, it becomes critical to assess in advance whether they are suitable for a specific target task.
\textit{Model transferability estimation} is an emerging and growing area of interest, aiming to propose a metric to quantify this suitability without training them individually, which is computationally prohibitive.
Despite extensive recent advances already devoted to this area, they have custom terminological definitions and experimental settings.
In this survey, we present the first review of existing advances in this area and categorize them into two separate realms: source-free model transferability estimation and source-dependent model transferability estimation. 
Each category is systematically defined, accompanied by a comprehensive taxonomy.
Besides, we address challenges and outline future research directions, intending to provide a comprehensive guide to aid researchers and practitioners.
A comprehensive list of model transferability estimation methods can be found at \url{https://github.com/YuheD/awesome-model-transferability-estimation}.
\end{abstract}

\begin{figure}[ht]
  \centering
  \includegraphics[width=.36\textwidth]{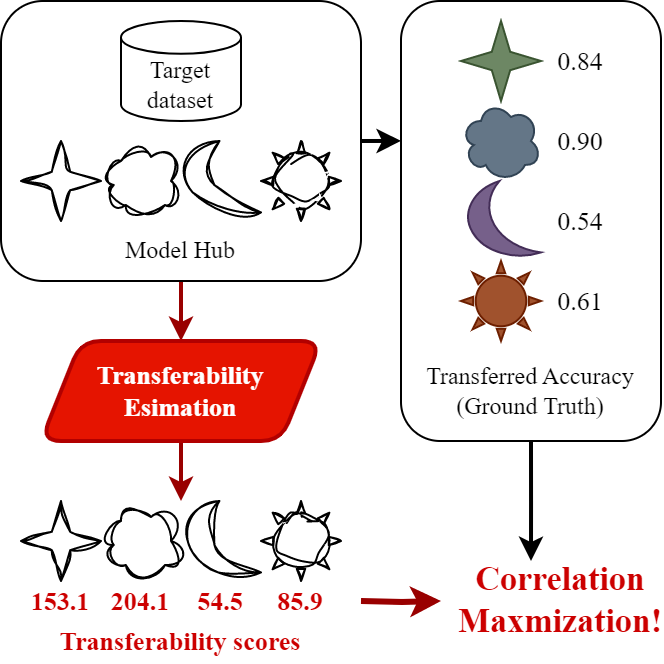}
  \caption{The paradigm of model transferability estimation.}
  \label{fig:overview}
\end{figure}

\begin{table*}[ht]
    \centering
    \caption{Comparison between model transferability estimation and some related learning paradigms.}
    \resizebox{.9\textwidth}{!}{
    \begin{tabular}{l|l|l}
        \toprule
        \multicolumn{1}{c|}{\textbf{Learning paradigm }} & \multicolumn{1}{c|}{\textbf{Input}} & \multicolumn{1}{c}{\textbf{Ground truth}} \\ \midrule
        Source-free model transferability estimation & Model hub, target set & Target task accuracy \\ \midrule
        Source-dependent model transferability estimation & Model hub, source set, target set & Target task accuracy \\ \midrule
        Task transferability estimation & Source task hub, target set & Target task accuracy \\ \midrule
        Generalization gap prediction & Model, training set & Difference between training and test accuracy \\ \midrule
        Out-of-distribution error prediction & Model, training set, test set with OOD samples & Test set error rate \\ \midrule
        Supervised validation & Training checkpoints, validation set & Test accuracy \\ \midrule
        Unsupervised validation & Training checkpoints, test set & Test accuracy \\ 

        \bottomrule
    \end{tabular}
    }
    \label{tab}
\end{table*}

\section{Introduction}
Transfer learning is a mature and widely applied field. 
It strives to improve the performance at the target task by transferring relevant yet different information from the source pre-trained models or datasets.
Such an approach can reduce the dependency of the target task on large-scale data and enhance the reuse of source data and models.
The pre-training and fine-tuning paradigm is the most representative paradigm, where the model is initially pre-trained on a source dataset and then fine-tuned for the target task. 
In recent years, a substantial number of pre-trained models have been available to the public, which have become one of the cornerstones of deep learning \cite{you2021logme}.

However, different target tasks benefit from distinct source models and datasets. 
Hence, given a specific target task, determining which one is the most suitable has become a key challenge. 
A simple solution is to brute-force train the target task on them individually, which is computationally prohibitive.
Model transferability estimation (MTE) aims to provide a metric to quantify this suitability efficiently.
An illustration of MTE is demonstrated in Fig. \ref{fig:overview}.
MTE methods predict a transferability score for each candidate model and these scores should be highly correlated with the transferred accuracy obtained after training through transfer learning algorithms.
Over recent years, MTE has achieved considerable advancements in fields like computer vision \cite{agostinelli2022stable} and neural language processing \cite{bai2023determine}.

In this paper, we present the first survey on MTE to introduce the recent advances in this field.
We divide the MTE studies into two primary realms, \ie, source-free model transferability estimation (SF-MTE) and source-dependent model transferability estimation (SD-MTE).
These two realms are distinguished based on whether they require access to the source dataset when evaluating transferability.
For each realm, we provide a comprehensive taxonomy from the perspective of the method and thoroughly review these advanced algorithms.

Our contributions can be summarized as follows,
\begin{itemize}
    \item To our knowledge, this survey is the first to systematically overview two distinct topics in transferability estimation: source-free model transferability estimation (SF-MTE, Sec. \ref{sec:mte}) and source-dependent model transferability estimation (SD-MTE, Sec. \ref{sec:dte}).
    \item We introduce a novel taxonomy of existing methods and provide a clear definition for each topic. We hope this survey will aid readers in developing a better understanding of the advancements in each topic.
    \item We provide an outlook of recent emerging trends and unresolved problems in Sec. \ref{sec:dscs}, offering insights into potential future research directions in the field of model transferability estimation.
\end{itemize}

\noindent
\textbf{Comparisons with previous surveys.}
A previous survey on transferability in deep learning \cite{jiang2022transferability}, which is highly relevant to our work, offers a comprehensive exploration of this concept across various domains of deep learning. 
This survey interlinks isolated areas within deep learning, illustrating their connection to transferability.
Their work can be viewed as a general overview of the entire field of transfer learning. In contrast, our work specifically focuses on the methods that estimate the transferability of source models.

\section{Related Research Topics}
We compare model transferability estimation (MTE) with several related topics, shown in Table \ref{tab}.

\noindent
\textbf{Task Transferability Estimation (TTE).}
Task transferability \cite{zamir2018taskonomy}, also known as task similarity, aims to explore the relationship among visual tasks and offers a principled approach to identify redundancies across tasks.
This topic is typically significant in multi-task learning and meta-learning problems.
In these problems, highly similar tasks can often be jointly learned, leading to a decrease in the need for annotation and improved performance.
Some TTE methods focus on finding the most suitable \textit{source task} from a data hub for a target task, while the model information is agnostic.
In contrast, MTE primarily focuses on the selection of models from a model hub, considering that the model's architecture, parameters, and training algorithms can all affect transferability. 

\noindent
\textbf{Generalization Gap Prediction.} 
Generalization Gap Prediction methods predict the difference between the accuracy of the training data and unseen test data from the same distribution, \ie, generalization gap.
The generalization gap represents the model's ability to generalize from the training data to new, unseen data, constituting a vital aspect in the evaluation and enhancement of machine learning models.
This task focuses on assessing a model's ability to generalize within the same task or domain, while model transferability estimation predicts the cross-task performance.

\noindent
\textbf{Out-of-distribution Error Prediction.} 
In practical application, the presence of Out-of-distribution (OOD) data presents substantial hurdles for deployed machine learning models, as even minor changes may result in considerable declines in performance.
OOD error prediction involves assessing the model's performance using a test set that includes OOD data.
It evaluates the generalization performance of a model within the same task across different data distributions, without involving transfer learning training for downstream tasks.

\noindent
\textbf{Validation.}
Validation is a critical step in the machine learning workflow, as it provides a way to assess and compare the performance of different training checkpoints. 
In the supervised validation setting, a supervised validation set is used for selecting the model with the best validation performance. 
On the other hand, unsupervised validation aims to leverage the unlabeled test set to assess these models.
Both types of validation tasks aim to achieve higher test performance, while MTE aims to obtain the best-transferred performance after downstream training.

\section{Source-Free Model Transferability Estimation (SF-MTE)} \label{sec:mte}

\begin{figure}
  \centering
  \includegraphics[width=0.5\textwidth]{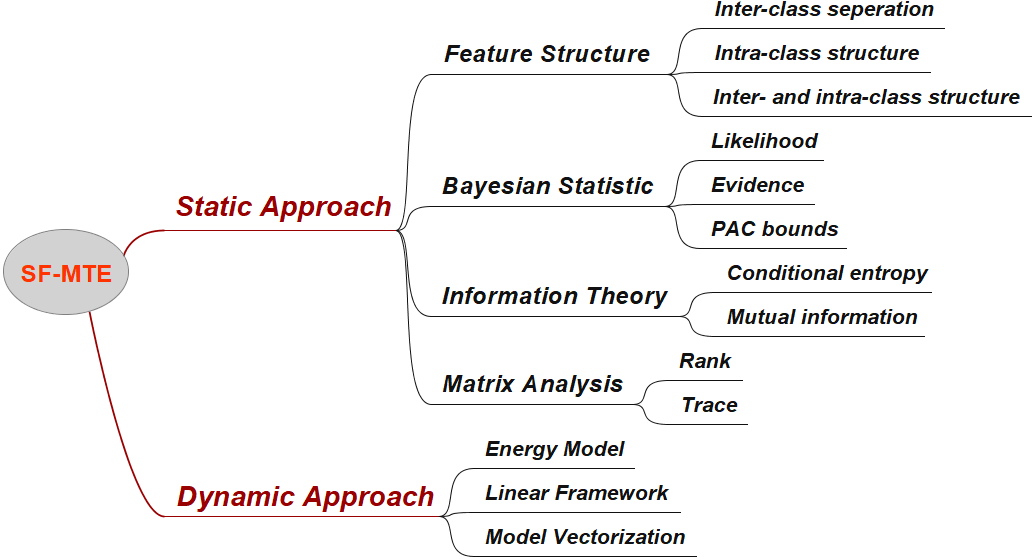}
  \caption{Taxonomy of source-free model transferability estimation methods (Sec. \ref{sec:mte}).}
  \label{fig:MT}
\end{figure}

In this section, we introduce existing source-free model transferability estimation (SF-MTE) methods in detail.
The overall taxonomy of SF-MTE is shown in Fig. \ref{fig:MT}. 
Specifically, we systematically categorize them into static and dynamic methods.
Static methods calculate scores directly based on model statistical information, while dynamic methods utilize some learning framework or representation space mapping algorithm to transform this statistical information before computing scores.
We start by introducing the problem setup and then delve into these two categories respectively.

\noindent
\textbf{Problem Setup.}
Considering an arbitrary target task on dataset $\mathcal{D}^T$ with a sample number of $n_t$.
Given a source pre-trained model hub $\mathcal{M} = \{\phi_i\}^{n_\mathcal{M}}_{i=1}$ with a total of $n_\mathcal{M}$ pre-trained models, the goal of SF-MTE is to predict a transferability score $s_i$ for each model $\phi_i$.
For an effective SF-MTE method, the correlation coefficient $C = Cor(\mathcal{S}, \mathcal{A})$ should be high, where $Cor(\cdot)$ denotes a correlation metric, $\mathcal{S}=\{s_i\}^{n_\mathcal{M}}_{i=1}$, $\mathcal{A} = \{Acc(\phi_i, \mathcal{D}_T)\}^{n_\mathcal{M}}_{i=1}$ is the ground truth, \ie, the transferred accuracy, and $Acc(\cdot)$ is the test accuracy of the target task after the training of transfer learning algorithm.

\subsection{Static SF-MTE}
Static methods refer to those that calculate scores directly based on some statistical information, such as features and logits computed on the target dataset using candidate source models.
Based on their techniques, we categorize existing static methods into four groups in this section, \ie, feature structure-based, Bayesian statistic-based, information theory-based, and matrix analysis-based methods.

\subsubsection{Feature Structure}
Methods in this group indicate that models exhibiting a beneficial structure tend to yield a higher transferability score.
Due to the simple output space, these methods are primarily used for handling classification tasks, primarily focusing on the intra-class and inter-class relationship of features.

\noindent
\textbf{Inter-class separation.}
If the features are class-wise separable in the feature space of the corresponding model, it is easier to find a decision boundary, allowing for good classification performance for the target task.
This is a natural idea, and numerous works are devoted to finding a metric to measure the separation.
A typical method is GBC \cite{pandy2022transferability}, which models the features in per-class Gaussian distributions with means and covariances of intra-class features. 
Then inter-class separation is measured by the summation of the Bhattacharyya coefficients between class pairs. 

H-score \cite{bao2019information} is another typical method, that calculates the matrix product between the pseudo reverse of the feature matrix and the covariance of the class means matrix. 
Subsequently, it calculates the trace to derive a transferability metric. 
A higher H-score indicates a larger inter-class variance and lower feature redundancy.
However, Ibrahim \etal \shortcite{ibrahim2022newer} observe that H-score becomes less accurate when calculated in high-dimensional feature spaces.
To address this issue, they propose a shrinkage estimator, named shrinkage-based H-score, to improve the robustness of the H-score on small target datasets. 

Separation Index (SI) \cite{kalhor2020ranking} is a distance-based complexity metric, which compares their labels to the nearest neighbor for each target sample. 
Specifically, samples with matching neighbor labels receive a score of 1, otherwise, the score is 0. 
SI is calculated by the summation of these scores for each sample in the target dataset.
In this way, a higher SI implies sufficient margin space between different classes.
Besides, since separate class features often lead to high accuracy in KNN (K-Nearest Neighbor) classification, CLID \cite{lu2023using}, a self-supervised method for evaluating pre-trained models, evaluates inter-class separation using KNN clustering learnability.

\noindent
\textbf{Intra-class structure.}
A fast and general method TMI \cite{xu2023fast} measures the transferability by the intra-class feature variance, which is a general concept widely used in metric learning.
Specifically, conditional entropy is employed for approximate estimation.
TMI considers transferability as a measure of generalization ability, which can be assessed with intra-class variance. 
This indicates how effectively the feature can be utilized for the target task.
Yang \etal \shortcite{yang2023pick} focus on the medical image  segmentation problem and propose a method named CC-FV.
CC-FV considers the intra-class consistency by using the distribution of features extracted from foreground voxels of the same class in each sample to calculate the distance.
Besides, they further measure the global feature diversity by the uniformity of feature distribution, which reflects the effectiveness of the features themselves.
Similar ideas are also evident in many other works \cite{bao2019information,ibrahim2022newer}.

\noindent
\textbf{Inter- and intra-class structure.}
Classical literature \cite{papyan2020prevalence} suggests that a more pronounced Neural Collapse (NC) in a well-trained model is indicative of better performance.
NC refers to the phenomenon where, in the terminal phase of training, features collapse to the class mean of the corresponding class, and the class means to align with the classifier's weight distribution at the vertices of a simplex equiangular tight frame (ETF).
Higher levels of NC should lead to features that exhibit both inter-class separation and intra-class compactness.
Many studies leverage the metric of NC and its variants as an indirect way to assess transferability.
Suresh \etal \shortcite{suresh2023pitfalls} directly analyze intra-class variability collapse and inter-class discriminative ability of the penultimate embedding space for the source model.
FaCe \cite{ding2023unleashing} also considers the feature variance collapse. 
Besides, the entropy of the similarity matrix as the metric of class fairness is taken into account, which can be seen as an approximation to the ETF structure.
Wang \etal \shortcite{wang2023far} consider both feature variability and the approximation to ETF, and they further explore the applicability of nearest-neighbor classifiers.

Besides, Ueno \etal \shortcite{ueno2020a} propose six basic metrics based on three hypotheses, including inter-class separability, intra-class feature diversity, and feature sparsity.
TMI \cite{xu2023fast} argues that intra-class variance should be larger, corresponding to stronger generalization.
This is a controversial issue that remains unresolved.

\subsubsection{Bayesian Statistic}
In this section, we introduce some methods based on the joint distribution of source and target. 
These methods can also be viewed as predictions of the source model's generalization ability on target data.
Likelihood is the primary perspective in the Bayesian-based methods.
The most well-known likelihood-based family of methods is LEEP \cite{nguyen2020leep} and the variants. 
LEEP is the average log-likelihood of the log-expected empirical predictor, which is a non-parameter classifier based on the joint distribution of the source and target distribution.
Several variants of LEEP have been developed for transferability estimation.
$\mathcal{N}$LEEP \cite{li2021ranking} simply improves LEEP by replacing the output layer of the models with a Gaussian mixture model.
In this way, the LEEP score can be applied to unsupervised or self-supervised pre-trained models that do not have a classification head.
Agostinelli \etal \shortcite{agostinelli2022transferability} design four metrics, \ie, MS-LEEP, E-LEEP, IoU-LEEP, and SoftIoU-LEEP, for variant settings.
MS-LEEP and E-LEEP use the log of mean predictions and the mean of log predictions respectively for the multi-source and source ensemble selection.
IoU-LEEP and SoftIoU-LEEP are designed specifically for the segmentation task.
Zhang \etal \shortcite{zhang2022efficient} leverage the similarity of model feature likelihood distributions for the semantic segmentation backbone evaluation problem.
SFDA \cite{shao2022not} is a dynamic method, that first maps the features into the proposed self-challenge space, then calculates the sum of log-likelihood as the metric.
Besides, Schreiber \etal \shortcite{schreiber2023model} utilize marginal likelihood for model selection in the wind and photovoltaic power forecasts problem.

PAC-Bayesian bound \cite{germain2016pac} is a framework that combines ideas from the PAC (Probably Approximately Correct) learning framework with Bayesian probability theory. 
It is used to analyze the generalization performance of Bayesian machine learning algorithms.
Ding \etal \shortcite{ding2022pactran} introduce this technique for MTE, yielding three different metrics with Dirichlet, Gamma, and Gaussian priors, respectively.
PACTran is a theoretically grounded framework that is particularly well-suited for target tasks involving classification.

\subsubsection{Information Theory}
Information theory involves the quantification of information.
Entropy and mutual information, two key elements in information theory, are leveraged for MTE.
Tran \etal \shortcite{tran2019transferability} treat training labels as random variables and introduce the negative conditional entropy (NCE) between labels of target samples extracted by a pre-trained model and the corresponding features.
They demonstrate that this value is related to the loss of the transferred model, thereby reflecting its transferability.
TransRate \cite{huang2022frustratingly} measures the mutual information between features of target samples extracted by a pre-trained model and the corresponding labels.
They use the code rate with a small distortion rate as the approximation of entropy and conditional entropy.

\subsubsection{Matrix Analysis}
Matrix analysis methods are leveraged to assess the feature matrix redundancy.
H-score \cite{bao2019information} and its variants also consider the informativeness of the target features provided by the models.
They directly calculate the trace of the covariance of the feature matrix.
RankMe \cite{garrido2023rankme}, proposed for self-supervised model evaluation, shares a similar idea.
They directly take the rank of the feature matrix as the metric.

\subsection{Dynamic SF-MTE}
Dynamic methods compute scores using mapping functions or learning frameworks to map the original static information into a different space. 
Dynamic methods usually consider the changes in the model during fine-tuning and aim to approximate this variation using an efficient transformation.

\subsubsection{Energy Model}
Energy models introduce a function that maps the input data to a single, non-probabilistic scalar called energy.
The energy models learn to make the energy functions assign lower values to the training data and higher values to others.
ETran \cite{gholami2023etran} introduces the energy score to quantify whether the target dataset is in-distribution or out-of-distribution for the candidate models.
For the same target dataset, a model that achieves a higher in-distribution level typically exhibits greater transferability.
PED \cite{li2023exploring} exclusively focuses on the concept of potential energy to model the feature dynamics during target task training.
They provide an energy-based perspective by treating the fine-tuning optimization process akin to a physical system seeking a state of minimal potential energy.
PED can be seen as feature remapping, and it can be combined with existing methods.

\subsubsection{Linear Framework}
Target task training can be simulated using an efficient linear framework, avoiding time-consuming computations and obtaining approximate results.
Kim \etal \shortcite{kim2016learning} train the Bayesian least square SVM (LS-SVM) classifier and select the best model by evidence of LS-SVM.
LogME \cite{you2021logme,you2022ranking} introduces a linear model upon target features and suggests estimating the maximum average log evidence of labels given the target features. 
LogME doesn't depend on a source classifier, making it a general approach that can be applied across various pre-trained model hubs, target tasks, and modalities.
TLogME \cite{fouquet2023transferability} applies LogME separately to the regression and classification aspects in detection, then combines them to create the final transferability score.
These marginal evidence-based methods are less prone to overfitting, which potentially improves its generalization ability.

SFDA \cite{shao2022not} simulates the fine-tuning procedure of transfer learning based on a regularized Fisher Discriminate Analysis and a self-challenging mechanism. 
EMMS \cite{meng2023foundation} solves this transferability estimation problem for multi-modal target tasks.
They employ large-scale foundation models to transform diverse modal labels into a unified label embedding space.
Subsequently, they develop a simple weighted linear square regression model to assess the transferability for various tasks.
Deshpande \etal \shortcite{deshpande2021linearized} analyze the generalization bounds and training speed by a Neural Tangent Kernel-based linear framework.
They propose two criteria based on the label-gradient correlation and label-feature correlation.
Li \etal \shortcite{li2023guided} propose a new perspective to re-frame this problem as a recommendation system.
Given a model hub, they consider using linear regression and FM (Factorization Machine) as the recommendation models.

\subsubsection{Model Vectorization}
Vectorizing the model or task to embedding is originally an idea from meta-learning \cite{achille2019task2vec}.
This approach maps the model and the target dataset to the same space and then measures the similarity between them in this unified space.
SynLearn \cite{ding2022pre} establishes a joint embedding space for models and target tasks, enabling the computation of the metric distances between them.
Model Spider \cite{zhang2023model} share a similar idea, which tokenizes the models and the target task by summarizing their characteristics into tokens and measures the transferability by their similarity.

\section{Source-Dependent Model Transferability Estimation (SD-MTE)} \label{sec:dte}
\begin{figure}
  \centering
  \includegraphics[width=0.47\textwidth]{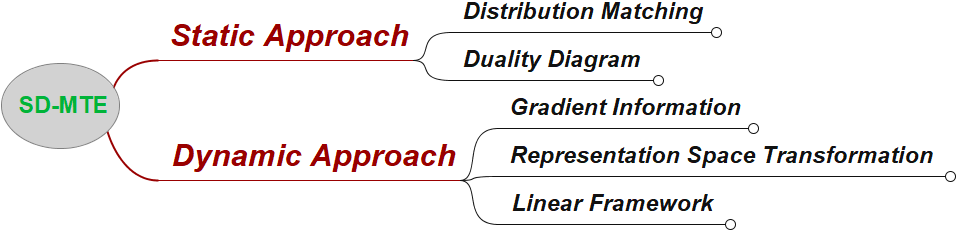}
  \caption{Taxonomy of source-dependent model transferability estimation methods (Sec. \ref{sec:dte}).}
  \label{fig:DT}
\end{figure}

In this section, we introduce the recent advances in source-dependent model transferability estimation (SD-MTE).
The taxonomy of SD-MTE is shown in Fig. \ref{fig:DT}.
The problem setup of SD-MTE is similar to SF-MTE introduced in Sec. \ref{sec:mte}.
The only difference is that the model hub in SF-MTE is a source-dependent version, \ie, $\mathcal{M}=\{\phi_i, \mathcal{D}^S_i\}_{i=1}^{n_\mathcal{M}}$.
Then we categorize these methods into static and dynamic approaches and introduce each category respectively.

\subsection{Static SD-MTE}
Since source data is available, it is possible to assess the domain gap using distribution distance metrics directly. 
The source dataset and the corresponding source model typically perform better on the target task when the gap is small.

\subsubsection{Distribution Matching}
Optimal transport (OT), also known as the Wasserstein distance (W-distance) or Kantorovich-Rubinstein norm, is a mathematical framework used to measure the minimum cost or distance required to transport a set of objects from one distribution to another while satisfying certain constraints. 
Tan \etal conduct extensive research on the application of OT in DTE and propose many efficient methods, including OTCE, JC-NCE, F-OTCE, and JC-OTCE.
OTCE \cite{tan2021otce} divides their transferability score into domain difference and task difference.
They solve the OT problem between source and target data to capture the underlying geometry of data and evaluate the domain difference.
OTCE is effective for classification tasks with sparse output space, thus the authors further extend OTCE to the semantic segmentation problem \cite{tan2021transferability} by sub-sampling pixels from input images.
However, a fundamental drawback of OTCE is the reliance on auxiliary tasks with known transfer performance to assist in determining the intrinsic parameters of the linear model.
JC-NCE \cite{tan2021practical} first solves the OT problem to obtain the joint probability distribution of source and target and then calculates the transferability score via negative conditional entropy. 
F-OTCE and JC-OTCE \cite{tan2022transferability} are proposed to dramatically improve the estimation efficiency without requiring auxiliary tasks.

Mehra \etal \shortcite{mehra2023analysis} introduce an upper bound, which consists of the W-distance between the transformed source and target distributions, the conditional entropy between the label distributions of the two tasks, and the weighted loss of the source classifier on the source task.
Subsequently, they put forward an optimization model aimed at minimizing the proposed upper bound to estimate transferability.
Additionally, based on the W-distance, WDJE \cite{zhan2023transfer} introduces a non-symmetric metric for object detection. 
Chen \etal \shortcite{chen2023estimate} propose a method addressing the problem of ranking pre-trained speech models. 
OSBORN \cite{bachu2023building} focused on model ensemble, specifically multi-source model selection.

Tong \etal \shortcite{tong2021mathematical} decouple the transferability with three key factors: the similarity between the source and target set, the sample sizes of the two, and the model complexity, characterized by the number of model parameters.
They propose a mathematical framework for transfer learning and establish a metric from the above three aspects, and $\chi^2$-distance is utilized for source and target distribution matching.

\subsubsection{Duality Diagram}
Duality Diagram Similarity (DDS) is a generic framework designed to represent and compare data with varying feature dimensions.
The duality diagram of a feature matrix $X\in\mathbb{R}^{n*d}$ is a triplet $(X,Q,D)$, where $Q\in\mathbb{R}^{d*d}$ quantifies the dependencies between individual feature dimensions and $D\in\mathbb{R}^{n*n}$ assigns weights for the images. 
Dwivedi \etal \cite{dwivedi2020duality} utilize DDS to express the features of the source and target feature matrix, and systematically compare different models and their layers based on a DDS similarity calculation method, facilitating a better understanding of the transferability of the models.
RSA \cite{dwivedi2019representation} can be formulated as a special case of DDS.
Specifically, RSA adopts techniques from computational neuroscience and computes similarity scores using Spearman's correlation on the lower triangular representation dissimilarity matrices between the source and target models.

\subsection{Dynamic SD-MTE}
Similar to dynamic SF-MTE, dynamic SD-MTE involves computing scores based on dynamic features such as gradients or using learning frameworks to map the original static information into a different space.

\subsubsection{Gradient Information}
The direction and magnitude of gradient updates reflect the dynamic characteristics of a sourced model.
Qi \etal \shortcite{qi2022transferability} consider the optimization procedure to be a distance approximation between the initial and the optimal points in the parameter space.
They calculate the principle gradient expectation (PGE) of the source and target dataset and compare their similarity to approximate the transferability.
Besides, attribution map matching is utilized by Song \etal \shortcite{song2019deep}.
They generate the attribution map by forward-and-backward propagation, and the gradient is a key component in the computation.

\subsubsection{Representation Space Transformation}
Wu \etal \shortcite{wu2021model} isolate the usage of source data from the target but still require leveraging source data to obtain a specification, \ie, reduced kernel mean embedding for each source.
They translate the input into the reduced kernel mean embedding, which can be regarded as a specification of the source, and compare the similarity between the candidate source and the target based on their specification.

Model2Vec \cite{achille2019task2vec} models the joint interaction between the source dataset and the source model to learn a joint embedding of the two. 
This embedding, denoted as $m$, is formulated as $m = F + b$, where $F$ represents the embedding of the source set, and $b$ is the model bias that introduces adjustments to the task embedding to account for the model-specific characteristics.
The dataset embedding $F$ is computed based on the estimation of the Fisher information matrix associated with the probe network parameters.
The transferability is measured by the distance between the model vector and target set embedding.

\subsubsection{Linear Framework}

Nguyen \etal \shortcite{nguyen2023transferability} address the challenge of estimating the transferability of regression models from source to target tasks. 
They introduce two linear estimators, namely Linear MSE and label MSE, to approximate the transferred target model using mean squared error.
Linear MSE estimates transferability by performing regression between features extracted from a model trained on the source model and the labels of the target set.
Label MSE, on the other hand, estimates transferability by conducting regression between the dummy labels obtained from the source model and the ground-truth labels of the target data.
Guo \etal \shortcite{guo2023identifying} introduce the reduced model as the specification in their learnware paradigm. 
They utilize cross-entropy and KL divergence to train a linear model and extract the parameters as the specification. 

\section{Evaluation and Study}
\subsection{Evaluation}

\noindent
\subsubsection{Metric}
An efficient MTE metric should predict a score for the candidate source model which is highly correlated with their transferred accuracy.
Thus correlation coefficient is usually utilized to quantify the effectiveness of the MTE method.
There are three common correlation coefficients.

\noindent
\textbf{Pearson correlation coefficient ($\rho$)} \cite{pearson1900mathematical} is a statistical measure that quantifies the strength and direction of the linear relationship between two continuous variables,
\begin{equation}
    \rho = \frac{\sum{(s_i - \bar{s})\cdot(a_i - \bar{a})}}{\sqrt{\sum{(s_i - \bar{s})^2}\cdot \sum{(a_i - \bar{a})^2}}}, 
\end{equation}
where $s_i$ and $a_i$ are the individual transferability score and transferred accuracy of i-th model, $\bar{s}$ and $\bar{a}$ are the means of $s$ and $a$, respectively.
$\rho$ measures linear correlation, and is sensitive to outliers. 

\noindent
\textbf{Kendall rank correlation coefficient ($\tau$)} \cite{fagin2003comparing} is a  non-parametric metric of correlation for ranked, discrete variables. 
$\tau$ doesn't care about the specific numerical values of scores, it focuses on evaluating the ranking or order of a model's transferability score within the entire model hub.
$\tau$ is formulated as,
\begin{equation}
    \tau = \frac{\sum_{i<j} \sigma(s_i - s_j) \cdot \sigma(a_i - a_j)}{\sqrt{\sum_{i<j} (\sigma(s_i - s_j))^2 \cdot \sum_{i<j} (\sigma(a_i - a_j))^2}},
\end{equation}
where $\sigma$ is the signal function. 
Compared to $\rho$, $\tau$ is less affected by outliers because it deals with the ranks of score points rather than their actual values.

\noindent
\textbf{Weighted Kendall rank correlation coefficient ($\tau_w$)} \cite{vigna2015weighted} is a variant of $\tau$ that takes into account the importance or weight assigned to each pair of transferability scores.
$\tau_w$ pays more attention to top-performing models, the formula is as follows,
\begin{equation}
    \tau_w = \frac{\sum_{i<j} w_{ij} \cdot \sigma(s_i - s_j) \cdot \sigma(a_i - a_j)}{\sqrt{\sum_{i<j} w_{ij} \cdot (\sigma(s_i - s_j))^2 \cdot \sum_{i<j} w_{ij} \cdot (\sigma(a_i - a_j))^2}},
\end{equation}
where $w_{ij}$ is the weight assigned to the pair of the transferability score and accuracy $(i,j)$.
A desired effective MTE method is expected to yield a positive coefficient close to 1. 

\subsubsection{Settings and Applications}
The benchmark in the MTE field has not yet been standardized. 
Experimental setups tend to vary for different methods. 
Here, we introduce some commonly used setups from the perspectives of the model hub and target task as follows. 

For the aspect of model hubs, the fundamental settings can be classified into several primary categories:
(i) \textit{Architecture selection}: When the source data used for training models in the model hub is identical, choose an appropriate model architecture.
(ii) \textit{Source selection}: In scenarios where the model architecture is fixed or agnostic, select a suitable source dataset.
(iii) \textit{Layer selection}: If the source model consists of multiple layers, determine the suitable layer to which a target task projection head should be appended.
(iv) \textit{Un-/self-supervised algorithm selection}: When the source models are pre-trained using un-supervised or self-supervised algorithms without classifiers or regressors, select an appropriate one.
(v) \textit{Hybrid model selection}: The four mentioned model hubs are the most common experimental setups, but researchers are often not limited to evaluating these four single-variable tasks alone. 
They typically consider the evaluation across a multitude of multi-variable model hubs, encompassing different architectures, source datasets, and pre-training algorithms, which is more practical. 

For the aspect of target datasets, most methods primarily focus on \textit{supervised} settings, where labels for the target dataset are available. 
This is especially essential for methods that rely on feature inter-class relationships and joint distributions between source and target.
However, these methods lose effectiveness when dealing with unsupervised target tasks.
Therefore, in recent years, there have been several advances \cite{garrido2023rankme,lu2023using} that do not rely on target dataset labels, referred to \textit{unsupervised} model transferability estimation.
These settings of model hubs and target datasets above can be applied to arbitrary tasks in machine learning, computer vision, or neural language processing society.
Existing works have focused on various target tasks, including image classification, semantic segmentation, text classification, medical image process, speech classification, \etal.

\subsection{Study}

Some prior research has conducted small-scale studies on transferability estimation.
Agostinelli \textit{et al}. \shortcite{agostinelli2022stable} conduct a study on the stability of existing MTE methods. 
They choose five representative methods H-score \cite{bao2019information}, GBC \cite{pandy2022transferability}, LogME \cite{you2021logme}, LEEP \cite{nguyen2020leep}, $\mathcal{N}$LEEP \cite{li2021ranking}, and conduct experiments on a broad range of 715k experimental setup variations, including source dataset selecting in semantic segmentation, model architecture selection in image classification, \etal
They observe that minor variations in experimental setups can result in differing conclusions regarding the superiority of one transferability metric over another.

Chaves \etal \shortcite{chaves2023performance} evaluate seven MTE methods, H-score, NCE \cite{tran2019transferability}, LEEP,  $\mathcal{N}$LEEP, LogME, GBC, Shrinkage-based H-score \cite{ibrahim2022newer}, in three medical applications, including out-of-distribution scenarios. 
Their study revealed that there are still no transferability scores capable of reliably and consistently estimating target performance in medical contexts.

Bassignana \etal \shortcite{bassignana2022evidence} and Bai \etal \shortcite{bai2023determine} explore the effectiveness of the existing MTE method in the neural language processing field.
Bassignana \etal \shortcite{bassignana2022evidence} conduct the encoder selection study in NLP on 10 diverse classification and structured prediction tasks.
Their findings suggest that quantitative metrics are more reliable than pure intuition and can assist in identifying unexpected Language Model candidates.
Bai \etal \shortcite{bai2023determine} conduct a detailed and comprehensive empirical study for the fourteen methods based on a multi-task neural language understanding benchmark.
They demonstrate the strengths and weaknesses of surveyed methods and show that H-score generally performs well with superiorities in effectiveness and efficiency.

\section{Discussion}
\label{sec:dscs}

\subsection{Emerging Trends}

\noindent
\textbf{Wider application areas.}
While much of the existing work in MTE has predominantly focused on traditional tasks such as image classification and semantic segmentation, an increasing number of MTE methods are now targeting object detection \cite{fouquet2023transferability}, text classification \cite{ding2023unleashing}, structured prediction \cite{bassignana2022evidence,bai2023determine}, and speech processing \cite{chen2023estimate}, as well as medical image processing \cite{yang2023pick}.

\noindent
\textbf{More complicated model hub.}
Existing work typically validates the effectiveness of single-variable model hubs, such as source datasets, model architecture, \etal 
In practice, models in model hubs often come from various sources, and there may be interactions between different variables. 
Recent works \cite{zhang2023model,ding2023unleashing} have started to explore mixed-model hubs that encompass multiple variables.

\noindent
\textbf{Less annotation.}
The majority of current research is conducted under supervised target dataset settings, where the predicted ground truth refers to the accuracy obtained after supervised transfer learning. 
However, there is a growing trend towards unsupervised or weakly supervised tasks, and recent advances have also addressed unsupervised scenarios \cite{garrido2023rankme}.

\subsection{Open Problems}

\noindent
\textbf{Robustness.}
Existing methods are sensitive to experimental settings, and even small variations between experimental setups can lead to different conclusions regarding the superiority of a metric compared to other metrics \cite{agostinelli2022stable}.
Exploring a stable and robust method is an unresolved problem.

\noindent
\textbf{Unified benchmarks.}
Currently, large-scale open-source unified benchmarks are still scarce. 
Each work conducts experiments on different model hubs, target tasks, and training parameters for obtaining transferred accuracy, leading to varying conclusions. 
Additionally, current research does not distinguish between source-free and source-dependent methods, which may be unfair to source-free methods.

\noindent
\textbf{Various transfer learning paradigms.}
\textit{Pre-training then fine-tuning} is the most commonly used transfer learning paradigm, and existing MTE methods largely rely on this paradigm to calculate ground truth accuracy. 
However, there is a wide variety of model-centric transfer learning methods such as source-free domain adaptation \cite{fang2022source}, test-time adaptation \cite{liang2023comprehensive}, \etal.
The open problem of how different transfer learning algorithms related to the transferability of models remains unexplored.

\noindent
\textbf{Foundation models.}
Recently, foundation models such as the GPT and CLIP family, have attracted much attention, while their transferability has not yet been explored. 
These models come with more parameters and higher training costs, making it an urgent issue to estimate the transferability.

\section{Conclusion}

Learning how to select an appropriate pre-trained model for downstream tasks is an emerging key problem in transfer learning. 
This survey provides a comprehensive survey of two related realms, \ie, source-free model transferability estimation and source-dependent model transferability estimation, which are unified under the broader learning paradigm of Model Transferability Estimation (MTE). 
For each realm, we introduce its definition and then present a new taxonomy of advanced approaches. 
We then introduce the evaluation metrics and application areas of MTE. 
Additionally, we provide insights into emerging research trends and open problems related to MTE. 
We believe this survey will facilitate both novices and experienced researchers in obtaining a better understanding of the current state of studies and challenges in the MTE field.

\footnotesize
\bibliographystyle{named}
\bibliography{ref}

\end{document}